\documentclass[conference, a4paper]{IEEEtran}
\IEEEoverridecommandlockouts

\usepackage[utf8]{inputenc}
\usepackage{comment}
\usepackage[lofdepth,lotdepth]{subfig}
\usepackage{array}
\usepackage{algorithm}
\usepackage{algorithmic}
\usepackage{cite}
\usepackage{amsmath,amssymb,amsfonts}
\usepackage{algorithmic}
\usepackage{graphicx}
\usepackage{diagbox}
\usepackage{color,soul}
\usepackage{xcolor}
\def\BibTeX{{\rm B\kern-.05em{\sc i\kern-.025em b}\kern-.08em
    T\kern-.1667em\lower.7ex\hbox{E}\kern-.125emX}}
    
\usepackage{graphicx}

\begin{document}

\title{Drift anticipation with forgetting to improve evolving fuzzy system
{\footnotesize \textsuperscript{}}
\thanks{}
}

\author{\IEEEauthorblockN{Clement Leroy}
\IEEEauthorblockA{\textit{Intuidoc Team} \\
\textit{Univ Rennes, CNRS, IRISA}\\
Rennes, France \\
clement.leroy@irisa.fr}
\and
\IEEEauthorblockN{Eric Anquetil}
\IEEEauthorblockA{\textit{Intuidoc Team} \\
\textit{Univ Rennes, CNRS, IRISA}\\
Rennes, France \\
eric.anquetil@irisa.fr}
\and
\IEEEauthorblockN{Nathalie Girard}
\IEEEauthorblockA{\textit{Intuidoc Team} \\
\textit{Univ Rennes, CNRS, IRISA}\\
Rennes, France \\
nathalie.girard@irisa.fr}
}

\maketitle

\begin{abstract}
Working with a non-stationary stream of data requires for the analysis system to evolve its model (the parameters as well as the structure) over time. In particular, concept drifts can occur, which makes it necessary to forget knowledge that has become obsolete. However, the forgetting is subjected to the stability-plasticity dilemma, that is, increasing forgetting improve reactivity of adapting to the new data while reducing the robustness of the system.
Based on a set of inference rules, Evolving Fuzzy Systems - EFS - have proven to be effective in solving the data stream learning problem.
However tackling the stability-plasticity dilemma is still an open question. 
This paper proposes a coherent method to integrate forgetting in Evolving Fuzzy System, based on the  recently introduced notion of concept drift anticipation.
The forgetting is applied with two methods: an exponential forgetting of the premise part and a deferred directional forgetting of the conclusion part of EFS to preserve the coherence between both parts.
The originality of the approach consists in applying the forgetting only in the anticipation module and in keeping the EFS (called principal system) learned without any forgetting.
Then, when a drift is detected in the stream, a selection mechanism is proposed to replace the obsolete parameters of the principal system with more suitable parameters of the anticipation module.
An evaluation of the proposed methods is carried out on benchmark online datasets, with a comparison with state-of-the-art online classifiers (Learn++.NSE, PENsemble, pclass) as well as with the original system using different forgetting strategies.

\end{abstract}

\begin{IEEEkeywords}
Evolving Fuzzy System EFS, Learning with Forgetting, Non Stationary data stream, Anticipation.
\end{IEEEkeywords}

\section{Introduction}


Data stream learning problem has become a new topic of interest that breaks with classical batch learning model for several reasons:
\begin{itemize}
    \item The learning algorithm must process one instance at a time without requiring access to previously seen data (One-shot learning).
    \item The stream is potentially infinite, thus instances should only be saved for a short time.
    \item Knowledge contained in the data stream can change over time, that is called concept drift.
\end{itemize}
In addition, most of the application using data stream have real-time constraints, requiring a fast processing
(take for example, the monitoring of network traffic and the credit fraud identification \cite{Babcock2002}, recommendation systems which take up the new recent context to propose more interesting content \cite{Linqi2016} or even a customized-command-gesture recognition system \cite{Manuel2017}).
To cope with theses new constraints, new incremental learning algorithms have been designed, inspired by classic batch approaches.  
For example, we can cite decision tree (CVFDT \cite{Hulten2001}, Hoeffding tree \cite{Ikonomovska2011}), or the ensemble classifiers (DWM \cite{Kolter2003}, Learn++.NSE \cite{Elwell2011}) or the Evolving Fuzzy Systems - EFS - (pclass \cite{Pratama2014}, ANYA \cite{Angelov2011}). 

For each of them, the adaptation of the model to the stream implies the incremental adaptation of the model parameters (such as the updating of sufficient statistics) and the evolution of the structure (addition/replacement of subtrees, classifier or fuzzy rules). Thus, the data stream learning problem can be simplified in two cases \textbf{if ... then ...}:
\begin{itemize}
    \item \textbf{First case: If} new data are close to the concepts already seen \textbf{Then} update the model parameters on these new data, perhaps using a forgetting factor in case of smooth drift (called incremental drift).
    \item \textbf{Second case: If} new data comes from the appearance of a new concept or a new class, \textbf{Then} update the model structure, perhaps using a forgetting factor in case of brutal drift.
\end{itemize}
The two main goals of a learning model are to guess in which case the system is after receiving new data, and what is the extent of the adaptation required.
Evolving fuzzy systems are suitable to address the data stream learning problem. They are granular models composed of fuzzy rules which locally adapt the distribution of points with a premise part, and discriminate the classes in a conclusion part. 
In the recently introduced ParaFIS\cite{Leroy2019}, a new learning model for evolving fuzzy system has been proposed. It considers two systems in parallel. The principal system is updated assuming the first case, \textit{i.e.} no drift. But at the same time, the second system - the anticipation module - presupposes the need for a structural update to tackle the second case. Then, with a posteriori information, the  model can decide which assumption was the right one and can update the principal system from the anticipation module if a structural update is necessary to have the most suitable structure. \\
However, questions still arise: when and how to forget data.
The first question concerns to the well-known stability-plasticity dilemma, which says that if a forgetting factor is applied continuously with a high magnitude, the system will be reactive to drift at all times but will be less stable (\textit{i.e.} less efficient in the long run term). Conversely, if none or a few forgetting factors are applied, then the system will be more efficient in stationary phase because it learns on more points, but it will be less reactive in the event of drift.
The second question concerns the ways of forgetting data, \textit{i.e.} what information, previously learned on a point, should be forgotten. Two approaches are possible, either forgetting all the past information whatever its relevance for the present (blind adaptation \cite{Kolter2003}), or it is the learning model which selects the information to forget. \\
This paper addresses these issues on two levels, from a general perspective in data stream, to a specific application in evolving fuzzy system.
To the best of our knowledge, no stable approach of forgetting in the conclusion part is proposed in the state of the art. The main difficulty is the wind-up problem \cite{Milek1995} which leads to the collapse of the conclusion part if a forgetting is applied continuously. To address this problem, we propose to take advantage of the anticipation module introduced in ParaFIS, to integrate forgetting in the conclusion part of the EFS. The application of forgetting only in the anticipation module makes it possible to respect the trade-off between stability and reactivity. Indeed, as long as the system does not detect any change in the data stream, no forgetting is necessary and the principal system adapts better to the stationary environment. But once a drift is detected, the anticipation module learned with forgetting, will update the principal system to be more reactive to the drift.
In addition, thanks to the granularity of the EFS, only the parameters affected by the drift will be updated.
The contributions of the paper are as follows:
\begin{itemize}
    \item Integration of forgetting in the conclusion part,
    \item Selection of conclusion parameters to update in case of drift.
\end{itemize}

The paper is organized as follows. The section \ref{sec:ParaFIS} recalls the ParaFIS model with the concept of anticipation and its current limits.
The section \ref{sec:contribution} begins with a discussion of the problem encountered with forgetting in conclusion part and presents our contribution (forgetting in the conclusion part of the anticipation module and method of selecting parameters during an update of the principal system). 
Section \ref{sec:expe} presents two experiments to evaluate the contribution step by step, with, at the end, a comparison with state-of-the-art data stream classifiers.

\section{ParaFIS Model}
\label{sec:ParaFIS}

Before detailing our contribution, this section recalls the ParaFIS system \cite{Leroy2019} on which it is based. 
The subsection \ref{subsection:RelatedWorks} is a brief overview of related works in the evolving fuzzy systems.
The subsection \ref{subsection:FISArchi} describes the architecture of a Fuzzy Inference System (FIS), the subsection \ref{subsection:ruleAdaptation} presents the learning step of such a system and the subsection \ref{subsection:anticipationModule} details the anticipation module. Last, the advantages and problems remaining in ParaFIS are discussed in subsection~\ref{subsection:ParaFisDiscussion}.

\subsection{Related works}
\label{subsection:RelatedWorks}

Since two decades, many fuzzy inference systems have been designed to be learned in an incremental manner, FlexFIS \cite{Lughofer2008}, eTs \cite{angelov2004}, ANYA \cite{Angelov2011}. Most of them are based on the Takagy-Sugeno fuzzy system composed of a set of fuzzy inference rules $R=\{r_i, 0\leq i \leq N\}$ with an antecedent part (also called premise), and a conclusion part. The difference between models lays in the choice of the structure and the choice of the criteria used to evolve the structure. The premise can be prototype with spherical shape \cite{Lughofer2008}, elliptical shape \cite{Manuel2013} or cloud \cite{Angelov2011}. Conclusion can have multiple input single ouput structure - MISO or multiple input multiple ouput structure - MIMO \cite{angelov2004}. The criteria used to add/remove new rule can be a distance-based criteria \cite{Lughofer2008}, a split condition based on the error and volume's rule \cite{Lughofer2018_split}, or density-based condition \cite{Angelov2011}.
\\
ParaFIS system \cite{Leroy2019} is built from the generalized evolving fuzzy system used in many papers such as in \cite{lemos2011,Lughofer2018}.
The model is described below.

\subsection{Model Architecture}
\label{subsection:FISArchi}

The architecture of ParaFIS is based on the Takagy-Sugeno fuzzy system.
The strength of such system is to combine the adaptation of the data distribution in the antecedent part (a generative model) with the discrimination of classes in the conclusion part to better fit the decision boundaries.
Each rule's antecedent is defined with a prototype that is set by a cluster with a center $\mu_i$ and a covariance matrix $A_i$. The rule's conclusion is defined with $c$ polynomial functions $l_i^j$ (for rule i, class j), $c$ being the number of classes. Finally, the structure of a rule $r_i$, is as follows:
\begin{equation}
\textbf{IF } \text{ \textbf{x} is close to $\mu_i$ } \textbf{   THEN   } \text{ $y_i^1=l_i^1(\textbf{x})$ .. $y_i^c=l_i^c(\textbf{x})$  } 
\label{eq:structRule}
\end{equation}
The degree of the polynomial function is set to $1$ with $\pi_{ik}^j$ the polynomial coefficients (see Eq. (\ref{eq:polfunction})). 
\begin{equation}
y_i^j=l^j_i(\textbf{x})= \pi_{i0}^j + \pi_{i1}^j x_1  + \pi_{i2}^j x_2 + .. + \pi_{in}^j x_n = \textbf{$\Pi_i^j$} \textbf{x}
\label{eq:polfunction}
\end{equation}
The membership of \textbf{x} to a rule $r_i$, denoted $\beta_i(\textbf{x})$, is given by a multivariate cauchy function of the mahalanobis distance from $x$ to $\mu_i$ (see Eq.(\ref{eq:RBF})).
\begin{equation}
\beta_i(\textbf{x})=\frac{1}{(1+(\textbf{x}-\textbf{$\mu$}_i)A_i^{-1}(\textbf{x}-\textbf{$\mu$}_i)^T)}
\label{eq:RBF}
\end{equation}
Finally, the predicted class for \textbf{x} is given by Eq. (\ref{eq:predictedClass1}),(\ref{eq:predictedClass2}).
\begin{align}
class(\textbf{x})&=y=argmax_j \ y^j(\textbf{x})
\label{eq:predictedClass1} \\
 y^j(\textbf{x})&=\sum_{i=1}^{N}\bar{\beta}_i(\textbf{x})y_i^j
\label{eq:predictedClass2} \\
\bar{\beta}_i(\textbf{x})&=\beta_i(\textbf{x}) / \sum_{l=1}^{N}\beta_l(\textbf{x})
\label{eq:predictedClass3} 
\end{align}

\begin{figure} 
\centering
\includegraphics[width=\linewidth]{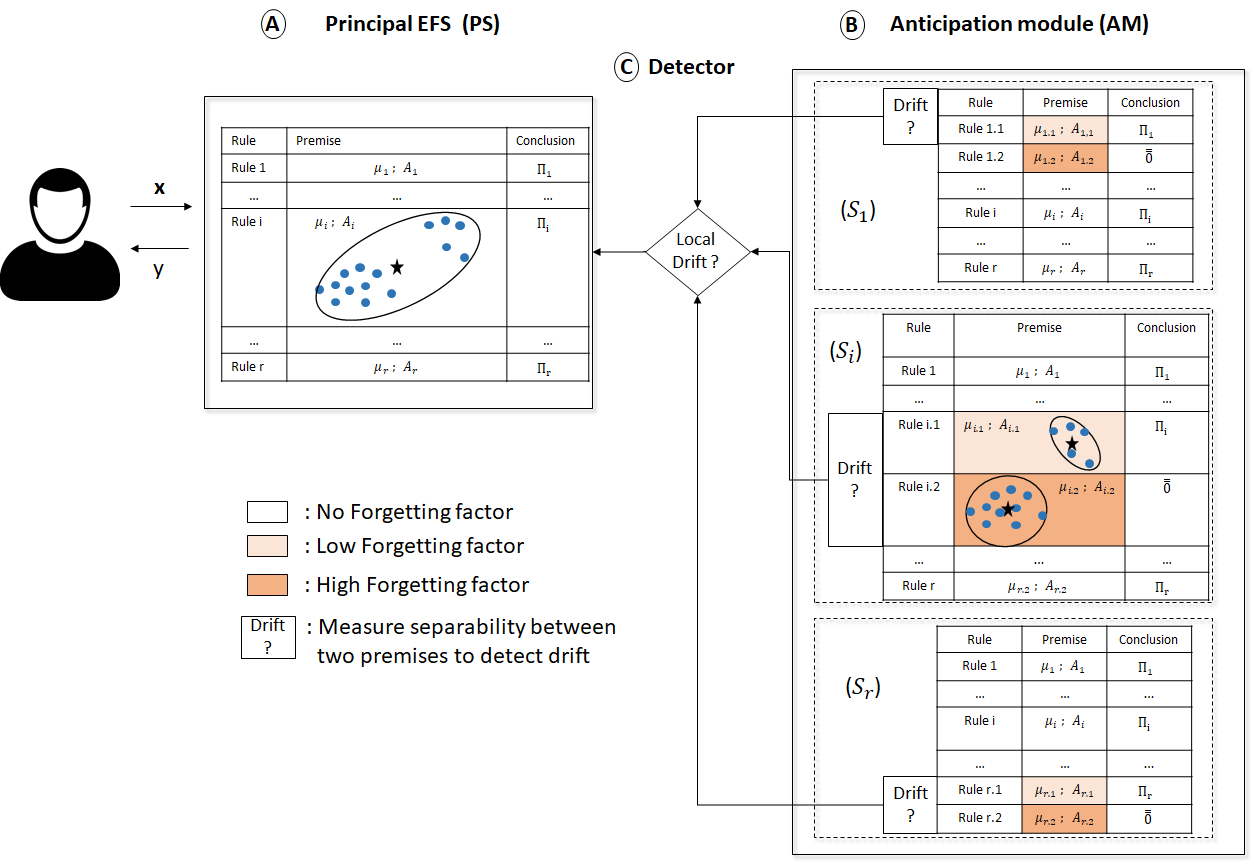}
\caption{Architecture of ParaFIS \cite{Leroy2019}}
\label{fig:paraFIS}
\end{figure}

The architecture of a FIS is illustrated in figure \ref{fig:paraFIS}, block (A).

\subsection{Rule's adaptation - Parameters adaptation}
\label{subsection:ruleAdaptation}
Each new incoming data $\textbf{x}_t$ is used to adapt the model parameters. In the premise part, only the most activated rule adapts its center and covariance matrix according to Eq. (\ref{eq:centerUpdate}),(\ref{eq:covUpdate}), in which, $\alpha=\frac{1}{t}$ is the fading factor where $t=min(k,tmax)$ (see \cite{Manuel2013}), and with $tmax$, the threshold defining the forgetting capacity, and $k$, the number of samples that activated the rules the most.
\begin{align}
\textbf{$\mu_t$}&= (1-\alpha) \mu_{t-1} + \alpha \textbf{x}_t
\label{eq:centerUpdate}
\\
\textbf{A$_t$}&= (1-\alpha) \textbf{A}_{t-1} + \alpha (\textbf{x}_t-\mu_t)(\textbf{x}_t-\mu_t)^T
\label{eq:covUpdate}
\end{align}
\\
The conclusion part is learned using a Weighted Recursive Least Square method (WRLS). In this optimisation problem, the weight - here the membership functions $\beta$ - are assumed to be almost constant to converge to the optimal solution. To reduce the computation time, the local learning of the conclusion part is often preferred \cite{Yen1998}. Thus, the rules are assumed to be independent to apply a RLS optimization on each one. 
The conclusion matrix $\Pi_{i(t)} = [\Pi_{i(t)}^1,..,\Pi_{i(t)}^c]$ of the rule $r_i$ at time $t$ (\textit{i.e.} after $t$ data points) is recursively computed according to:
\begin{align}
\Pi_{i(t)} &=\Pi_{i(t-1)} + C_{i(t)} \beta_i(\textbf{x})C_i \textbf{x} (Y_t-\textbf{x}\Pi_{i(t-1)})  
\label{eq:consUpdate}
\\
\text{Where } C_{i(t)} &= C_{i(t-1)}-\frac{\beta_i(\textbf{x}) C_{i(t-1)} \textbf{x} \textbf{x}^T C_i}{1+\beta_i(\textbf{x})\textbf{x}^T C_i \textbf{x} }
\label{eq:corrUpdate}
\end{align}
With $C_i$ a correlation matrix initialized by $C_{i(t=0)}=\Omega Id$ where $Id$ is the identity matrix and $\Omega$ a constant often fixed to $100$ (see \cite{Almaksour2011},\cite{angelov2004}).


\subsection{Anticipation Module}
\label{subsection:anticipationModule}
In ParaFIS, as shown in figure \ref{fig:paraFIS} - block (B), an anticipation module (AM) is added to the fuzzy inference system (called in this case principal system - PS). The goal of the anticipation module is to foresee an occurrence of a drift near each rule premise by anticipating the need of a structural update. Indeed, if a drift occurs in the vicinity of a rule, the distribution of point changes and a single rule is no longer sufficient to model the data. The idea of the anticipation is to consider for each rule $i$ of the FIS, an anticipated system $S_i$ where the rule $i$ is represented by two sub-rules $i.1,i.2$ to model the same distribution of points. In this ways, if a drift occurs, the anticipated system will already be effective in adapting to the drift before even detect it.
To do this, the anticipation module is learned in parallel with the principal system. Only the anticipated system ($S_i$) of the most activated rule $i$ is learned synchronously with the rule $i$.

In ParaFIS, the premise of the principal system does not have forgetting (There is no $tmax$)  while the premise of the two sub-rules $i.1,i.2$ from ($S_i$) have two different fading factors $\alpha_1$,$\alpha_2$. One sub-rule is learned with a low forgetting factor to capture information over a long period of time for stability, while the other is learned with a high forgetting factor to capture the most recent information in order to be reactive in case of drift. 

In addition, the detection of change in the distribution is done also in the anticipation module. The ParaFIS system detects that a rule $i$ is no longer sufficient to model the data (\textit{i.e.} a drift has occurred) if the premises of the two sub-rules $i.1,\ i.2$ in the anticipation system $(S_i)$ are sufficiently separated according to a clustering separability criterion given in eq. \ref{eq:Condition1}-\ref{eq:Condition2}. 
\begin{align}
\textbf{Condition 1}& \hspace{1.2cm} &  ||\mu_i-\mu_j|| &> k_s (\sigma_i + \sigma_j) &\hspace{1cm} 
\label{eq:Condition1} \\
\textbf{Condition 2}&  & k_i&> n_{min} &
\label{eq:Condition2}
\end{align}
Where $\sigma_i$ (resp. $\sigma_j$) is the distance between $\mu_i$ (resp. $\mu_j$) and the hyper-ellipsoid's envelope of the cluster $i$ (resp. $j$), along $(\mu_i,\mu_j)$ axis. $k_s$ is a coefficient related to the separation between cluster.
When the conditions are met for the two sub-rules $i.1,\ i.2$, the principal system is replaced by the anticipated system ($S_i$).

\subsection{Discussion}
\label{subsection:ParaFisDiscussion}
The ParaFIS system has two main advantages. First of all, the rule creation condition (\textit{i.e.} the detector) is based on a separability criterion between the clusters which is enough robust to noise contrary to distance-based conditions. Second, the anticipation of the premise part allows to maintain an efficient and stable principal system in the absence of drift, but still reactive and better fitted when a drift occurs, thanks to the structural update carried out from the anticipation module.
However, in ParaFIS as in any fuzzy inference system, the conclusion part has no forgetting capacity, which raises to two concerns. First, the system could be more reactive and more efficient after a drift if the conclusion were learned with forgetting. Second, forget only the premise part but not the conclusion part leads to an inconsistency in the system. Indeed, they are learned with different information if forgetting is applied differently, which could be damaging.

\section{Our Contribution}
\label{sec:contribution}
The paper's contribution is divided into three parts.
The first part is a discussion on the difficulty of introducing forgetting in the conclusion part. The second part is our proposal to integrate forgetting in the conclusion part of the anticipation module. The third part proposes two strategies to replace the conclusion of the principal system from the anticipation module, when a drift is detected.

\subsection{Discussion on forgetting and rules' conclusion}

The conclusion part of an evolving fuzzy system aims to discriminate between classes.
Its learning assumes that the underlying distribution does not evolve over time.
Typically, the $\beta(x)$ function is assumed constant over time, which is no longer true when drifts occur.
To maintain an efficient and accurate discrimination between classes over time and maintain consistency between the premise part and the consequent part, it is necessary to introduce the forgetting capacity in the conclusion.
The common approach is to exponentially weight the data over time (\cite{FWRLS}).
However, introduce forgetting in the RLS learning method without introduce instability is still an open question.
Indeed, if old data are forgotten whatever their significance, as in classical methods, then it causes the unbounded growth of the estimator (known as estimator blowup or covariance windup problem) leading to noise sensitivity and numerical difficulties \cite{Milek1995}.
Several ad-hoc approaches have been proposed, mainly based on regularization method with assumption or by introducing upper bound \cite{Milek1995}.
To the best of our knowledge, these methods are still not used in evolving fuzzy systems as they do not prevent collapse of the conclusion matrix over a long time. 
Recently in \cite{Manuel2013}, a new forgetting method, called the differed directional forgetting (DDF), has been proposed to forget the RLS parameters. The main idea is based on the concept of "directional forgetting", \textit{i.e.} to limit the windup phenomenon by directing the forgetting in the most excited dimensions in a sliding window. This offers a good compromise between forgetting data and maintaining the stability of the parameters.
However, as discussed earlier, the stability-plasticity dilemma tells us that it is damaging to forget data when there is no change in the data distribution. 

Starting from these problems, the next part introduces our contribution with a strategy to anticipate the forgetting in the conclusion part in order to maintain consistency with the premise part and the current data stream in the ParaFIS model.
\begin{figure}[t]
    \centering
    \includegraphics[width=\linewidth]{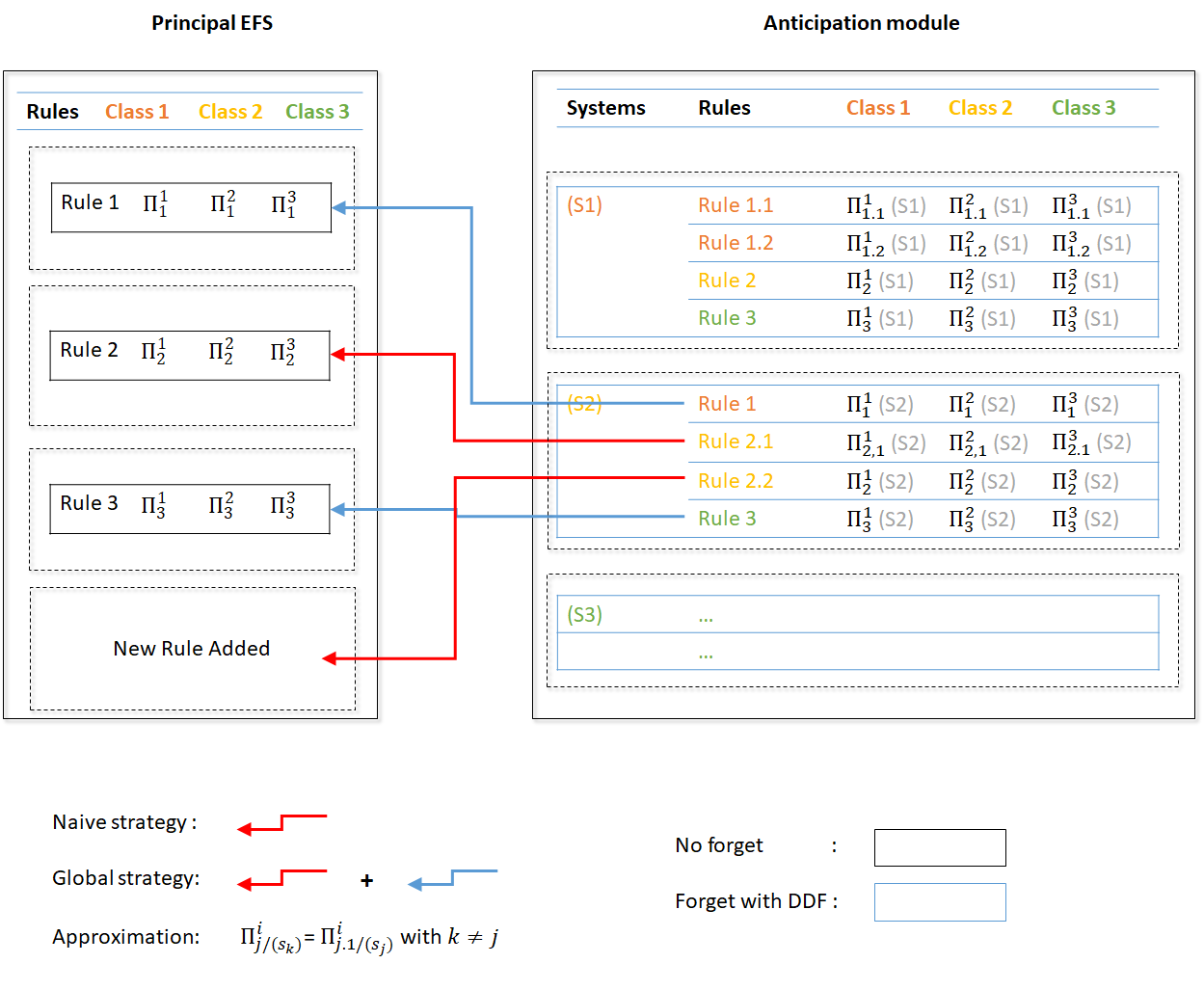}
    \caption{ Illustration of the two replacements strategies in the event that a drift is detected for rule 2. In the principal system, the 3 rules are represented with their respective conclusion. In the anticipation module, the 3 anticipated systems are represented with their own conclusions matrices learned with DDF. Replacement strategies are illustrated by the arrows.}
    \label{fig:replacementStrategy}
\end{figure}

\subsection{Anticipation principle and forgetting in conclusion part}

To integrate forgetting in the conclusion part, we propose to use the DDF method \cite{Manuel2013}.
In DDF, the data point are saved in a sequential sliding window of fixed size. Once the window is full, the oldest data point is used to recursively decrement the correlation matrix $C_i(t)$. The correlation matrix represents the directional forgetting matrix that is used to update the RLS parameters in the conclusions matrix $\Pi_{(t)}$. In this way, the correlation matrix is learned only on the data point from time $s$ to time $t$ with $t-s$ the window size. The decremental equation is given in eq.\ref{eq:DDF}, where $C_{i(s\rightarrow t)}$ is the correlation matrix of the rule $i$ learned on the data points from $s$ to $t$. 
On the contrary, the conclusion matrix is not decremented to preserve robustness of the consequent part and avoid windup problems. 
\begin{equation}
C_{i(s+1\rightarrow t)} = C_{i(s\rightarrow t)}+\frac{\beta_i(\textbf{x}_s) C_{i(s)} \textbf{x}_s \textbf{x}_s^T C_{i(s)}}{1+\beta_i(\textbf{x}_s)\textbf{x}_s^T C_{i(s)} \textbf{x}_s }
\label{eq:DDF}
\end{equation}

However, there are few pitfalls to avoid. Unlike to premise part where the center and the covariance matrix are computed for each rule independently of the other rules, the learning of the conclusion matrix of a rule depends on the others throughout the normalized $\Bar{\beta}(x)$ function. Indeed, if a drift occurs nearby the premise part of the rule $i$, then the conclusion part of all rules in the system will be impacted throughout $\Bar{\beta}(x)$.
As a result, the conclusion part of a rule cannot be anticipated without taking into account the other rules.

To compute the normalized beta for the anticipated conclusions, it is necessary to virtually built the "r" anticipated systems $S_i$ as illustrated in Figure \ref{fig:paraFIS} - block (B),  with r the numbers of rules. The "r" different assumptions lead to "r" different scalar fields of the normalized membership function used to compute the RLS parameters which will lead to r different conclusion matrices. As an example, let's consider the RLS update in the hypothesis where the local distribution fitted by the premise of rule 1 has drifted ($ S_1 $ system). In $S_1$, all conclusion matrices $\Pi_j, j\in[1,r]$ are updated using the last sample $x_t$ according to eq. \ref{eq:consUpdate}. The normalized membership function $\Bar{\beta}_i(x)$ for the rule $i$ is $\Bar{\beta}_i(x)=\frac{\beta_i(x)}{\sum_j \beta_j(x)} \ with \ j\in{1.1,1.2,2,3,..,r}$.
At end, each of the r anticipated systems requires the update of (r+1)*C hyperplanes. 
Updating the r*(r+1)*C hyperplanes on each point is time-consuming, so, it does not satisfy the real-time constraint.
The following section presents two strategies to reduce the complexity of the system.

\subsection{Strategies to update conclusion part from anticipation}

To decrease the computation complexity, the naive idea of considering only the conclusion matrices of the sub-rules $i.1,\ i.2$ regardless the others can be explored. In this naive approach, only the conclusions $\Pi_{i.1},\Pi_{i.2}$ are computed for the $(S_i)$ system ($i\in [1,r]$). Once a drift is detected for a rule $i$, it is replaced by the two sub-rules $r_{i.1},r_{i.2}$ without replacing the conclusions matrices of the other rules. This naive strategy is illustrated in Figure \ref{fig:replacementStrategy}.

The naive approach assumes that a drift in the local area of a rule will have no impact on the conclusions of the others. However, all rules make a decision to classify a point for any class. If a drift occurs for one class, then all rules' conclusions will be impacted. 

An other assumption can be done. The hyperplanes $\Pi_{j/(S_k)}$ of rules $j\neq k$ in the virtually built "$k$" anticipated system will be assumed identical and equals to $\Pi_{j.1/(S_j)}$. In this ways, for each of the anticipated system, all the conclusion matrices are known. Then, once a drift is detected nearby the premise of the rule $i$, the conclusion matrices of all rules are replaced by the conclusions matrices of the anticipated system as illustrated in Figure \ref{fig:replacementStrategy}. In this assumption, the other rules have also been learned with forgetting in the premise and conclusion part. It assumes also that the second sub-rules in each anticipated system will not impact too much the hyperplanes (\textit{i.e.} consider the rule $j\neq i$ is the same as consider rule $j.1$ and rule $j.2$ ).




\begin{table*}[t]

\begin{tabular}{  l  p{1.5cm} | p{1cm} | p{1.2cm} | p{0.7cm} | p{0.7cm} | p{0.7cm} | p{0.7cm} | p{0.7cm} | p{0.7cm} | p{0.7cm} | p{0.7cm} | p{0.5cm}  }
	Model & & Electricity Pricing & Hyperplane & Iris+ & Car & 10dplane & Weather & Sea & SinH & Line & Sin & Mean \\  \hline
	ParaFIS & No Forget & \textbf{77$\pm$15} & 91$\pm$03 & 82$\pm$14 & 8$\pm$11 & 68$\pm$34 & 78$\pm$03 & 94$\pm$04 & 67$\pm$09 & 85$\pm$15 & 85$\pm$13 & 81 \\
	& Forget PS & \textbf{77$\pm$15} & \textbf{93$\pm$02} & \textbf{85$\pm$12} & 79$\pm$12 & 63$\pm$16 & 78$\pm$03 & 97$\pm$01 & 71$\pm$07 & 93$\pm$06 & \textbf{94$\pm$06} & 83 \\
	& Forget AM Naive & \textbf{77$\pm$15} &\textbf{ 93$\pm$02} & 82$\pm$14 & \textbf{82$\pm$09} & 70$\pm$31 & 79$\pm$03 & 96$\pm$03 & 70$\pm$07 &92$\pm$10 & 93$\pm$09 & 83 \\
	& Forget AM Global & \textbf{77$\pm$15} \hspace{0.2cm}  &\textbf{ 93$\pm$02} \hspace{0.1cm}  & \textbf{85$\pm$15} & 81$\pm$10 & 77$\pm$34 & 78$\pm$03 & \textbf{98$\pm$01}  & 70$\pm$07  & \textbf{94$\pm$06} & \textbf{94$\pm$08} & \textbf{85} \\ \hline
	Learn++ & CDE  & 69$\pm$08 & 90$\pm$00 & \textbf{85$\pm$14} & 68$\pm$30 & 71$\pm$13 & 73$\pm$02 & 93$\pm$02 & \textbf{75$\pm$50} & 89$\pm$14 & 80$\pm$13 & 79 \\ 
	Learn++ & NSE  & 69$\pm$08 & 91$\pm$02 & 84$\pm$17 & 67$\pm$30 & 72$\pm$14 & 75$\pm$03 & 93$\pm$02 & 73$\pm$22 & 88$\pm$13 & 80$\pm$15 & 79 \\ 
	pENsemble & AxisParallel & 75$\pm$16 & 92$\pm$02 & 78$\pm$15 & 79$\pm$10 & 78$\pm$20 & \textbf{80$\pm$02} & 97$\pm$02 & 71$\pm$06 & 90$\pm$07 & 78$\pm$26 & 82 \\ 
	pENsemble & Multivariate & 75$\pm$16 & 92$\pm$02 & 75$\pm$17 & 79$\pm$10 & \textbf{80$\pm$20} & 78$\pm$02 & 97$\pm$02 & 71$\pm$06 & 90$\pm$07 & 78$\pm$30 & 82 \\ 
	pClass & & 68$\pm$10 & 91$\pm$02 & 73$\pm$18 & 77$\pm$10 & 63$\pm$26 & 68$\pm$04 & 89$\pm$10 & 71$\pm$09 & 91$\pm$07 & 72$\pm$20 & 76 \\ 
\end{tabular}
\caption{ \centering Final Results - Mean Accuracy Score }
\label{tab:FinalScore}

\end{table*}

\begin{figure}[h]
    \centering
    \includegraphics[width=\linewidth]{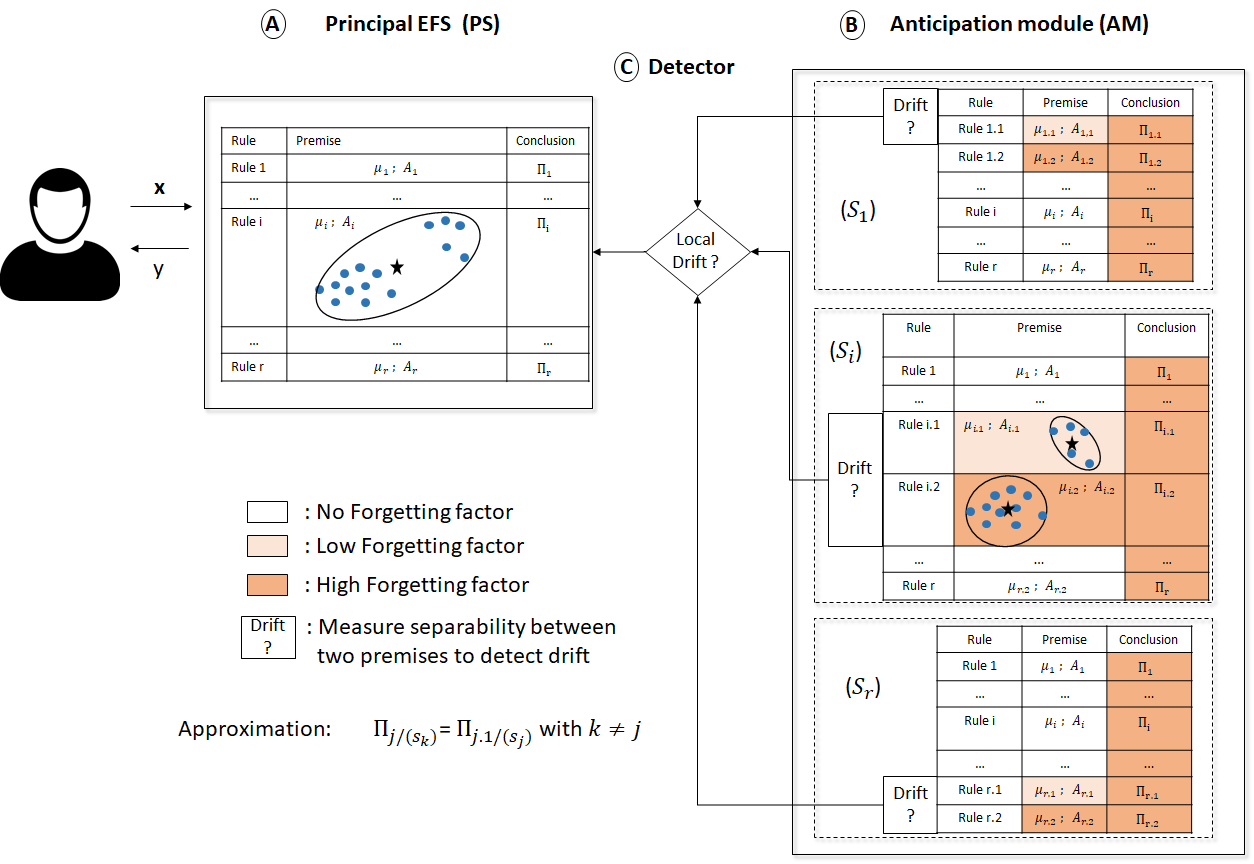}
    \caption{Final system obtained from ParaFIS with the addition of forgetting in the conclusion part of the anticipation module.}
    \label{fig:paraFIS+}
\end{figure}

At end, the final system with the contribution can be illustrated with the figure \ref{fig:paraFIS+}. The block (A) is a classical FIS that receives the data and gives the recognition label. The block (B) contains 'r' anticipated systems which are learned in parallel with forgetting (premise part + conclusion part). The block (C) is the drift detector based on a separability criteria applied on the premise part of two sub-rules of each anticipated system. Once a drift is detected in ($S_i$), the principal system is replaced by the anticipated one.

An important point is that the choice of evolving fuzzy system used as principal system is free. The choice done in the paper may not be the best one and for example an evolving fuzzy system with anticipated cloud structure \cite{Angelov2011} as premise could be better. But, the main suggestion is that using the anticipation concept and forgetting in conclusion can help any fuzzy system to be more reactive in case of drift while keeping stability in the other case, as it is with our choice of fuzzy system.

\section{Experiments}
\label{sec:expe}
Experiments are conducted to evaluate the advantages of forgetting the conclusion part with anticipation. 
The first and the second section introduce the benchmark dataset and protocol taken from the recent paper \cite{Lughofer2018}. The third section compares both, the naive and the global strategies, to update the conclusion part of the principal system after a drift. The fourth section compares the different strategies to apply forgetting in the conclusion part between [No forgetting, forgetting only in the anticipation module (Forget AM Naive/Global), forgetting continuously in the principal system (Forget PS) ]. The last section compares our contribution with state-of-the-art Evolving Fuzzy System and ensemble classifiers obtained from \cite{Lughofer2018} and shows improvements on 8 among 10 datasets containing different kind of drifts. The final mean accuracy scores are given in table~\ref{tab:FinalScore}.

\subsection{Evaluation protocol}

Evaluate the performance of a streaming algorithm requires different protocols from those used for evaluating classical learning algorithms. Many of them are discussed in \cite{Gama2013}. This paper is only concerned with the classification performance of a system. The simulation follows the periodic hold-out process where the stream of data is generated chunk by chunk. One chunk is used to train the system and then one chunk is used to test the system in an online mode. Thus, the system is evaluated every two chunks to built performance criteria over time. 
The performance is measured using the mean accuracy score and the standard deviation computed on all the chunks.
However, the existence of drift in the dataset naturally induces important fluctuation of the score independently of the classifier. To compensate for this, McNemar's significance test is also presented \cite{Dietterich1998}. The McNemar test is used to compare two classifiers evaluated only once over the same dataset \cite{Dietterich1998} as it is in our case. It consists in computing the statistic K given equation \ref{eq:mcnemmar} with $n_{0,1}$ the examples misclassified by the first classifier and not by the second and, $n_{1,0}$ the examples well-classified by the first classifier and not by the second. 
The K distribution converges to a $\chi^2$ distribution of degree 1. The null hypothesis of getting non significant difference between the two classifiers is rejected with a confidence score $\alpha$ if K is greater than a given threshold. A statistic K that rejects the null hypothesis with a confidence greater than $99\%$ ($K>6.63$) is noted by a \textcolor{green}{+}, between [90\%,99\%] by a \textcolor{gray}{$\approx$} and below $90\%$ ($K<2.7$) by a \textcolor{red}{-}.  If the numbers of contingent errors $n_{1,0}$+$n_{0,1}$ is below the recommended value of 25 to converge toward a Chi-square distribution, a \textcolor{orange}{(x)} is added.
The McNemar test is apply on the classifier using the best strategy "Forget AM Global" and the others strategy to measure a significance difference between both strategies. The proposed test can not be extended to the state-of-the-art benchmark results get from \cite{Lughofer2018} due to the unavailability of the classifiers (the classification score over each data point get from benchmark classifier is required).
\begin{equation}
\label{eq:mcnemmar}
    K=\frac{(n_{1,0}-n_{0,1})^2}{(n_{1,0}+n_{0,1})}
\end{equation}

\subsection{Datasets}

In order to evaluate the algorithm over different types of drifts (incremental/brutal/gradual/recurrent), the datasets sin, sinH, 10dplane, line, Car+, Iris+ have been chosen. They are generated with simulated drifts often using mathematical equation described in \cite{Minku2010}. The SEA dataset \cite{Kim2001}, in its extended version \cite{Polikar2013}, proposes to mix several types of drift with noise and imbalanced data. In addition, Weather dataset from \cite{Polikar2011} with incremental drifts is studied. The Hyperplanes obtained from supplemental material of paper \cite{Lughofer2018}, generated from the MOA frameworks \cite{MOAFramework} and the real world dataset Electricity pricing \cite{Harries99} are also investigated.
Table \ref{tab:dataset} gives the information on the datasets and presents the test parameters, with in the columns: IA: Input Attributes, C: Classes, DP: Data Points, TS: Time Stamps, TRS: Training Samples,TES: Testing Samples. Thus, all of these datasets cover a wide variety of data streams with different shapes of data distributions and different drifts.


\begin{table}[h]
    \centering
    \begin{tabular}{lcccccc}
        \hline 
        \hline
        Data stream & IA & C & DP & TS & TRS & TES \\
        \hline 
         SEA & 3 &2 & 100 000 & 200 & 250 & 250 \\
         Weather & 4 & 2 & 18159  & 400 & 30 & 30 \\
         Line & 2 & 2 & 2500 & 10 & 200 & 50 \\
         Sin  & 2 & 2 & 2500 & 10 & 200 & 50 \\
         Sinh & 2 & 2 & 2500 & 10 & 200 & 50 \\
         10dplane & 10 & 2 & 1200 & 10 & 100 & 20 \\
         Iris+& 4 & 4 & 450 & 10 & 34 & 11\\
         Car & 6 & 2 & 1728 & 10 & 130 & 42\\
         Hyperplane & 4 & 2 & 120K & 96 & 1000 & 250  \\
         Electricity pricing & 8 & 2 & 45312 & 199 & 150 & 77 \\
         \hline 
         \hline
    \end{tabular}
    \caption{Datasets description}
    \label{tab:dataset}
\end{table}

The paraFIS model contains 4 parameters to define:
\begin{itemize}
    \item $\alpha_1$,$\alpha_2$: the forgetting factor of sub-rules i1, i2; 
    \item $k_s$: the separation measure between cluster; 
    \item $ws$: the windows size for DDF. 
\end{itemize}
The $\alpha_1,\alpha_2$ parameters will depend on the features space dimensions. $\alpha_1$ is fixed to 200 and $\alpha_2$ to 10 or 30 (The best score is chosen). 
The $k_s$ parameters depend not only on the dimension space but also on the choice of $\alpha_1$, $\alpha_2$ and the data distribution. There is no rule to define it, so several values are tested between $[0.4,1]$ on a validation dataset (20\% of the total dataset) and the value that created rules in a "good" proportion is chosen. 
The window size of the DDF method depends on the dimensions of the space but also on the number of classes. To set it, all window size values are tested over a range of $[10,200]$ and the window with the best score is chosen.
The set of parameters of a dataset is the same for all the configurations tested (No forget,Forget PS,Forget AM naive,Forget AM global).

\begin{figure*}[h]
    \begin{center}
    \subfloat[Sea]{
      \includegraphics[width=0.30\linewidth]{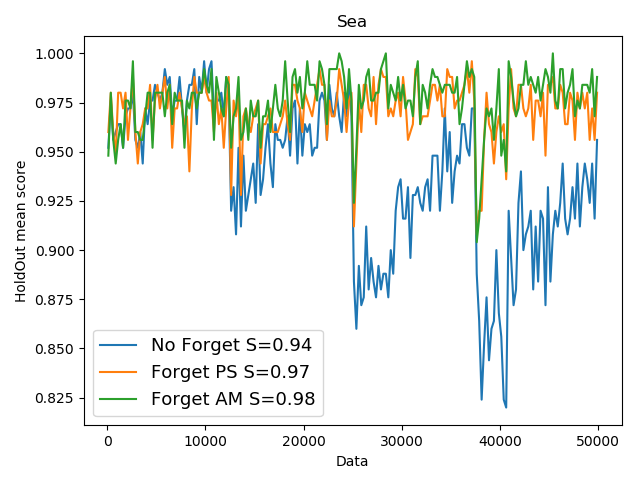}
         \label{fig:SEA}
     }
     \hfill
    \subfloat[Hyperplane]{
         \includegraphics[width=0.30\linewidth]{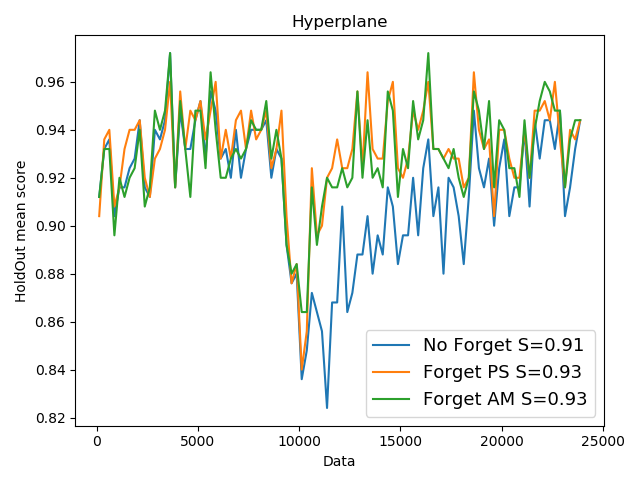}
         \label{fig:Hyperplane}
     }
     \hfill
    \subfloat[10dplane]{
         \includegraphics[width=0.30\linewidth]{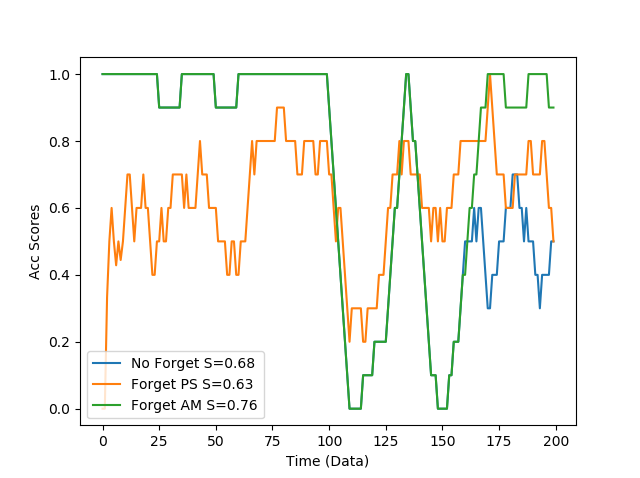}
         \label{fig:10dplane}
     }
        \caption{Example of holdout test with three datasets for the three strategies (No forget, forget in principal system PS, forget in anticipation module AM). S is the mean score.  }
        \label{fig:Comparison}
         
    \end{center}
\end{figure*}

\subsection{Comparison of Naive and Global update strategies}

In the previous part, two updating strategies have been proposed: the naive approach and the global replacement of the conclusion matrix with assumption.
These are two strategies used to satisfy real-time constraints based on handcraft assumptions. The first considers that it is more important to maintain the stability of rules far from the drift even if reactivity of adaptation of their hyperplanes is reduced. On the contrary, the second assumes that damaging the stability of the others rules is preferable to improve reactivity of hyperplanes when a drift occurs.
Both strategies are tested on all datasets, and the results of the models [Forget AM naive, Forget AM global] are presented in Table \ref{tab:FinalScore}. We can see that replacing all conclusion parts of all rules from the anticipation module is often a better solution. This means that the use of hyperplanes learned with directional forgetting, just after a drift, makes it possible to react better to the drift without damaging stability of old knowledge too much. This can be explained by the fact that the conclusion matrix is not decremented, only the correlation matrix is, to guide the learning. Thus, the old knowledge is still contained in the conclusion matrices and, after the switch, the decremented directional matrices will help the system to react more quickly to the drift by directing the conclusion learning on the new concept.

\subsection{Different strategies to apply forgetting in the conclusion part}

In order to measure the interest to apply forgetting with DDF only in the anticipation module, two others strategies are tested: the "No Forget" strategy and the "Forget Principal System" strategy (Forget PS). The first one is just the ParaFIS system described in Sec. \ref{sec:ParaFIS} where no forgetting capacity is applied in the consequent part. In the second one, forgetting is applied in the anticipation module as described in our contribution (Sec. \ref{sec:contribution}). In addition, forgetting is also applied in the principal system to get a system that continuously forgets.
When a drift is detected and the conclusions are replaced from the anticipation module to the principal system, the windows used in the anticipation module will also replace the one used in the principal system. This keeps the consistency of the memorized normalized activation in the DDF-windows with the conclusions matrix learning. Results of the Hold-Out test on all the datasets are presented in Table \ref{tab:FinalScore} in rows [ParaFIS No Forget, Forget PS, Forget AM Global].
Examples of plots of mean score by chunk are given in Figure \ref{fig:Comparison} for the Sea, Hyperplane and 10dplane datasets.
By comparing the rows "No Forget" and "Forget PS", we see that there is no dominant strategy. Sometimes it is preferable to forget the conclusion, sometimes not, depending on the stability-plasticity dilemma. However our strategy "Forget AM Global" allows a trade off between the two strategies by forgetting just at the time of the drift, which results in a better accuracy score on all datasets except for sinH.  To validate the significance of the results, a McNemar test is carried out by comparing the best strategy "Forget AM Global", with all the other strategies. Results are presented in the Table \ref{tab:testMcNemar}. We can see that for Electricity Pricing, Iris+, Car and SinH datasets, there is no difference in the proportion of error. For most of these datasets, the number of contingent errors is less than the recommended value of 25 due to the small size of the dataset. For 10dplane, SEA, and Line, where the difference in the mean accuracy score is the most important, there is a significant difference in the proportions of errors. The strategy "Forget AM Global" is significantly better for most of the datasets where a difference of mean accuracy score is observed in the Table \ref{tab:FinalScore}.

\begin{table}[t]
    \centering
\begin{tabular}{  p{1.5cm}  p{1.2cm} | p{1.2cm} | p{1.2cm}  }
    & \multicolumn{3}{c}{Strategies (compared to Forget AM Global)} \\ \\
	Dataset &  No Forget & Forget PS & Forget AM Naive  \\  \hline
	 Electricity Pricing &  \textcolor{red}{-} & \textcolor{red}{-} & \textcolor{red}{-} \\
	 Hyperplane &   \textcolor{green}{+}  & \textcolor{gray}{$\approx$}& \textcolor{gray}{$\approx$} \\
	 Iris+ &  \textcolor{red}{-} \textcolor{orange}{(x)} & \textcolor{red}{-} \textcolor{orange}{(x)} & \textcolor{red}{-} \textcolor{orange}{(x)} \\
	 Car &  \textcolor{red}{-} & \textcolor{red}{-} & \textcolor{red}{-} \textcolor{orange}{(x)}\\
	10dplane  &  \textcolor{green}{+} \textcolor{orange}{(x)}& \textcolor{green}{+} &\textcolor{green}{+} \textcolor{orange}{(x)}\\
	Weather &  \textcolor{green}{+} & \textcolor{green}{+} &\textcolor{green}{+} \\
	Sea &  \textcolor{green}{+} & \textcolor{green}{+} & \textcolor{green}{+} \\
	SinH & \textcolor{gray}{$\approx$} \textcolor{orange}{(x)}& \textcolor{red}{-} \textcolor{orange}{(x)}& \textcolor{red}{-} \\
	Line & \textcolor{green}{+} & \textcolor{red}{-} \textcolor{orange}{(x)}& \textcolor{green}{+} \\
	Sin &  \textcolor{green}{+} \textcolor{orange}{(x)}& \textcolor{red}{-} \textcolor{orange}{(x)}& \textcolor{gray}{$\approx$} \\
\end{tabular}
\caption{McNemar test between "Forget AM Global" and the other strategies - A statistic K that reject the null hypothesis with a confident above $99\%$ ($K>6.63$) is noticed by a \textcolor{green}{+}, between [90\%,99\%] by a \textcolor{gray}{$\approx$} and below $90\%$ ($K<2.7$) by a \textcolor{red}{-}. If the number of contingency errors is below the recommended value of 25, a \textcolor{orange}{(x)} is added}
\label{tab:testMcNemar}

\end{table}

\subsection{Comparison with state-of-the-art}

Finally, to validate the proposed method, a comparison with state-of-the-art streaming classifiers is carried out based on the results published in \cite{Lughofer2018}. Results are shown Table \ref{tab:FinalScore}. The pclass classifier \cite{Lughofer2015} is an evolving fuzzy system based on a generalized TSK fuzzy inference system as ParaFIS, with a rule creation process based on a recursive "density" function and with an online feature selection. Its ensemble version, namely pENsemble \cite{Lughofer2018}, is an ensemble classifiers using pclass as base learned. It is combined with an online drift detection, ensemble pruning and online features scenarios. Two other ensembles classifiers are compared: Learn++.NSE \cite{Polikar2011}, Learn++.CDE \cite{Polikar2013}. They are designed to deal with non stationary streams thanks to a dynamic voting scenario which reflects the current data streams. 
We can see in the results, that for 6 among the 10 benchmark datasets, the ParaFIS system outperforms state-of-the-art models. The average mean scores over all datasets have reached 85\% of accuracy score against 82\% for pENsemble, the best from state-of-the-art.

\section{Conclusion \& Outlooks}

This paper introduced a new strategy to include forgetting capacity in the conclusion part of an evolving fuzzy system, based on the deferred directional forgetting. The strategy relies on the anticipation concept recently introduced in the ParaFIS system. The results of the paper highlight that the consequent part of the EFS plays an important interdependent role in the classification performance of such system. Consequently, it is necessary to integrate forgetting to adapt the conclusion part to non stationary stream. The proposed method consists to update consequent part of all rules to the drift when it is detected. The updates is based on the anticipation of the forgetting of the conclusions part. It gives a convincing trade-off to the plasticity-stability dilemma. The performances obtained by the proposed approach, for a classification task, are superior to those of the state-of-the-art approaches, this for most of the tested datasets. However, these systems still have the thorny problem of the parameters setting. Indeed, how to set the parameters in a data stream context (with no data available) is a important question to be resolved, in particular with ParaFIS system which has 4 parameters. Future work will investigate how to define or adapt these parameters with the data stream.


\bibliographystyle{unsrt} 
\bibliography{biblio}

\end{document}